\documentclass[letterpaper, 10 pt, conference]{ieeeconf}  %

\IEEEoverridecommandlockouts                              %

\usepackage{times}
\usepackage{graphicx}
\graphicspath{ {images/} } %
\usepackage{multicol}
\usepackage[bookmarks=true]{hyperref} 
\usepackage[ruled,vlined,noresetcount ]{algorithm2e}
\usepackage{subcaption}
\usepackage{booktabs}
\usepackage[usenames,dvipsnames]{xcolor}
\usepackage[utf8]{inputenc}
\usepackage{amssymb}
\usepackage{amsmath}
\usepackage{commath}
\usepackage[nocompress]{cite}
\usepackage{ textcomp }
\def\BibTeX{{\rm B\kern-.05em{\sc i\kern-.025em b}\kern-.08em
    T\kern-.1667em\lower.7ex\hbox{E}\kern-.125emX}}

\begin{document}
\title{\LARGE \bf
Smart-Inspect: Micro Scale Localization and Classification of Smartphone Glass Defects for Industrial Automation}

\author{M Usman Maqbool Bhutta$^{1}$, Shoaib Aslam$^{2}$, Peng Yun$^{3}$, Jianhao Jiao$^{1}$ and Ming Liu$^{1}$$^{3}$%
\thanks{*This work was supported by the National Natural Science Foundation of China, under grant No. U1713211, the Research Grant Council of Hong Kong SAR Government, China, under Project No. 11210017, No. 21202816, and the Shenzhen Science, Technology and Innovation Commission (SZSTI) under grant JCYJ20160428154842603, awarded to Prof. Ming Liu.}%
\thanks{$^{1}$ Department of Electronic and Computer Engineering, HKUST, HK.
        {\tt\small mumbhutta@connect.ust.hk}}%
\thanks{$^{2}$ Department of Mechanical and Aerospace Engineering, HKUST, HK.
       }%
\thanks{$^{3}$ Department of Computer Science and Engineering, HKUST, HK.
       }%
}
\maketitle
\begin{abstract}
The presence of any type of defect on the glass screen of smart devices has a great impact on their quality. We present a robust semi-supervised learning framework for intelligent micro-scaled localization and classification of defects on a 16K pixel image of smartphone glass. Our model features the efficient recognition and labeling of three types of defects: scratches, light leakage due to cracks, and pits. Our method also differentiates between the defects and light reflections due to dust particles and sensor regions, which are classified as non-defect areas. We use a partially labeled dataset to achieve high robustness and excellent classification of defect and non-defect areas as compared to principal components analysis (PCA), multi-resolution and information-fusion-based algorithms. In addition, we incorporated two classifiers at different stages of our inspection framework for labeling and refining the unlabeled defects. We successfully enhanced the inspection depth-limit up to 5 microns. The experimental results show that our method outperforms manual inspection in testing the quality of glass screen samples by identifying defects on samples that have been marked as good by human inspection.
\end{abstract}

\IEEEpeerreviewmaketitle
\section{Introduction}
In the era of robotics and automation, artificial intelligence (AI) is helping to solve many difficult problems on a level at which humans are unable to reach. Glass inspection is one of the key challenging problem for the glass manufacturing industry. With the rise in smart device manufacturing and increased industrial competition, manufacturing companies are facing financial losses due to manual glass inspection. Carried out by a limited human workforce, manual inspection is costly, time-consuming and inconsistent. Furthermore, there may exist defects that the human eye cannot detect, compromising the quality of the product at a consumer level. Nowadays, companies are showing great interest in investing in automation systems along with state-of-the-art techniques, which can help them overcome these problems, thus boosting the production line and in return the sales profits.\\

A variety of systems have been proposed to solve the inspection problems for different market niches of limited types of glass. For defect inspection of satin glass and float glass, researchers have used machine learning techniques \cite{adamo2009online, peng2008online, adamo2010calibration, adamo2009low}. Several frameworks based on image-processing have been proposed for satin glass \cite{adamo2009low} and glass bottles \cite{adamo2008automated}. Optical-based approaches have been proposed to detect micro-cracks in glass \cite{sakata2017development} and surface defects in touch panel glass \cite{chang2016development}, and inspect window glass \cite{savolainen1995novel}. Contributions have been made related to rough set theory to defect detection of automotive glass for vehicles \cite{lee2002application}. Di Li et al. \cite{li2014defect} propose a method for surface defects inspection of smartphone cover glass. The authors applied a PCA algorithm to work on smartphone cover glass, whereas all the previous works using this method concentrate on LCD or general glass. For smartphone glass inspection, a deep learning-based approach is introduced in \cite{go2019deep}. \\

\begin{figure}
\includegraphics[width=\linewidth]{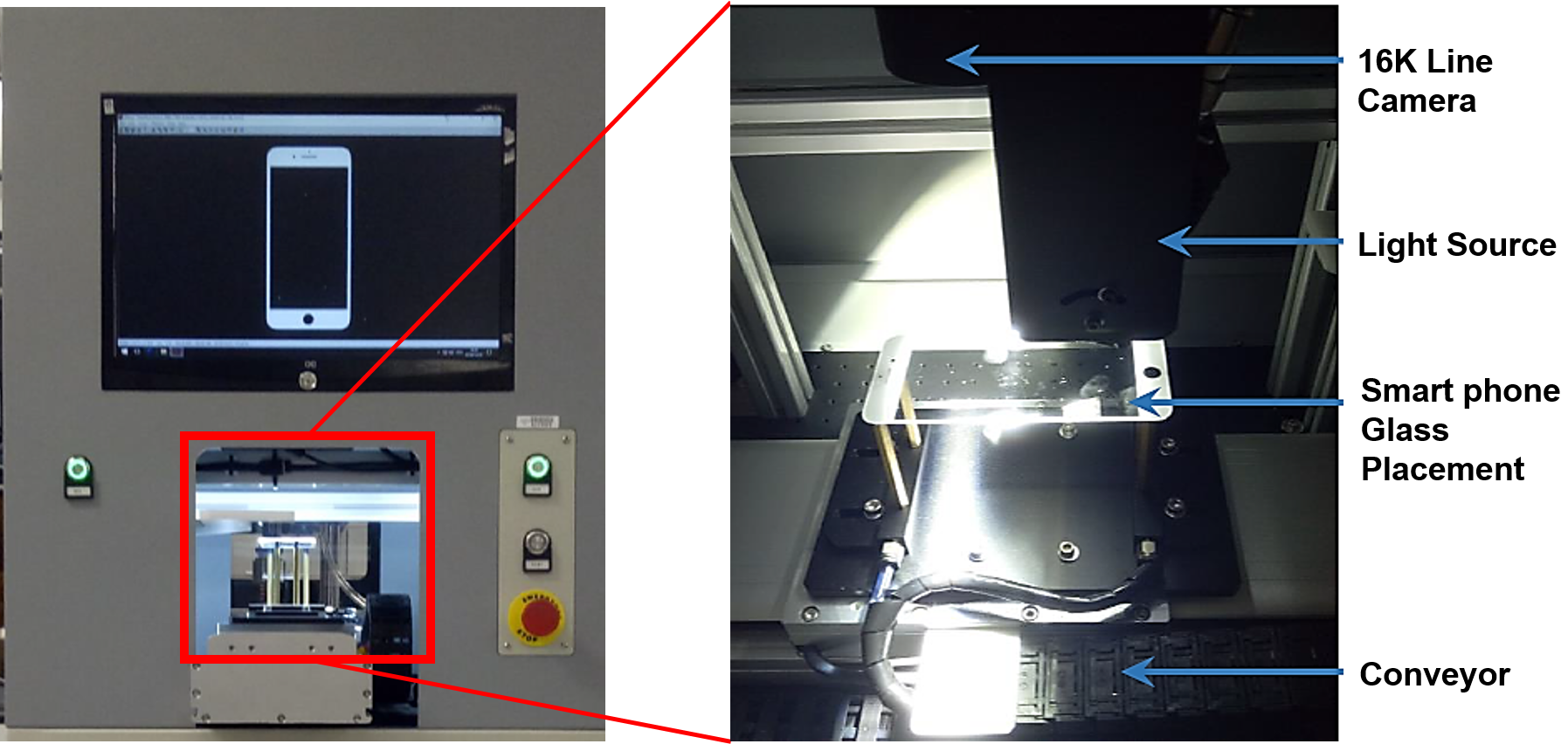}
\centering
\caption{Smartphone glass inspection system. The left image shows the front view, and the right image shows the inside view of the experimental system.}
\label{fig1}
\end{figure}
The production of smart device glass has experienced a huge rise in the last ten years. Technology companies who manufacture smart gadgets such as mobile phones, tablets, laptops, and smartwatches produce millions of sets each year. This high volume of production means that the inspection of smart device glass, not only requires a large manual workforce but also makes time another key constraint. Yet, the average time for manually inspecting the glass of one smart phone is about 1-2 minutes. Furthermore, the human eye is only able to detect defects over 0.1 millimeters, which limits the accuracy of the inspection. Therefore, meeting the immense market demand by increasing the production rate is becoming challenging without incorporating robotics, automation and AI to the production lines to solve these inspection problems.\\

For smartphone glass inspection, high-level accuracy and speed are the key-challenging tasks. Currently, state-of-the-art works \cite{li2014defect} present a mechanism based on principal components analysis (PCA) for defect inspection of smartphone cover glass. This is limited to detecting defects only. Using this technique, it is complicated to classify light leakages and specks of dust on the smartphone glass.\\

Our proposed scheme \textit{Smart-Inspect} uses an experimental setup for smartphone glass inspection, which will be discussed in detail in section \ref{Sec.Formulation and Methodolody}. Based on the dataset collected using the inspection system, we precisely localize the defects over a full-screen image of the glass and classify them to different types of defects such as scratches, pits and cracks, along with non-defect areas such as sensor regions, light reflections due to dust. The very large-sized image can have many combinations of defects and non-defect areas. Smart-Inspect makes the identification of defects, non-defect areas, and sensors regions possible, and characterizes the type of defects. \\

We begin by outlining related work in Sec \ref{sec.RelatedWork}, and present our problem description in Sec \ref{Sec.Problem Description}. Sec \ref{Sec.Formulation and Methodolody} presents our proposed method and framework for the smart inspection of smartphone screen glass. Sec \ref{Sec:Experiments} describes the experiments and presents the results obtained by Smart-Inspect, including both quantitative and qualitative evaluations. In Sec \ref{sec:conclusion} we conclude this work.
\section{Related Work}
\label{sec.RelatedWork}
This section presents a review of recent work related to glass inspection techniques and frameworks. In the literature, minimal contributions to this area specifically relating to smart device glass inspection techniques have been found. However, some methods are present that are related to detecting several glass defects in LCD and general glass. \cite{shimizu2000detection} introduces a fan-beam laser-light-based method for inspecting the scratches and dust over LCD panels. A method for the recognition of bubbles in the glass has been done in \cite{zhao2011method}. In this method, the authors propose a technique called binary feature histogram (BFH), which helps in the characterization and classification of glass defects.\\
An online distributed float glass inspection scheme is introduced in \cite{peng2008online}, which uses the OTSU method along with image filtration using gradient direction. The authors also use an adaptive surface for estimating the downward threshold. This system can quickly detect the bubbles, light reflection, and lards. \cite{perng2011automated} shows a two-phase method for an LED glass defect inspection framework using machine learning, which involves both a training and testing process. Research work based on wavelet analysis and fuzzy k-nearest neighbor is presented in \cite{liu2011classification} for the inspection of general glass. This approach can help to identify defects such as bubbles, inclusions, distortion, tin drop, and cracks. \\
The manufacturing process of mobile phone glass is very different from that of general glass. It requires a higher quality glass and therefore a higher level of diagnosis of the defects. For inspecting smartphone glass cover, \cite{li2014defect} developed a framework based on PCA in facial recognition to classify the defects. Each image is taken as a defect face and sets up a training set for defects features estimation and classification using PCA. The proposed system by \cite{li2014defect} not only detects the defects but is also able to recognize each one to some extent. This PCA-based method can help to identify typical defects such as scratches, cracks, edges, angle cutting, and deformation.\\
Semi-supervised learning methods exist such as Pseudo-Label \cite{lee2013pseudo}, learning using deep generative models \cite{kingma2014semi} and with ladder networks\cite{rasmus2015semi}, and learning by association \cite{haeusser2017learning}; however, these approaches require a large labeled dataset for the robustness evaluation. \\
A smartphone glass sample has many defects, along with sensor regions, which include holes for the camera, speaker, and buttons areas. The methods discussed above are not able to process a full glass image because they are unable to distinguish between the non-defect region and the original defects. Furthermore, the labeling of 16K glass image datasets for defects of down to 5 microns is also another challenging aspect, which cannot be handled by the previously discussed approaches. \\
\section{Problem Description}
\label{Sec.Problem Description}
Defects inspection and classification is the fundamental challenging problem. The methods discussed in 
\ref{sec.RelatedWork} show good results if the target is only to detect defects; however, they can perform poorly in classification. For example, these techniques are unable to distinguish defects such as scratches from dust, which is not a defect. A more critical challenging problem is the identification of a non-defect area sample because glass sample may contain written text or QR Codes. A further challenge is detecting defects that are very small in dimension.\\
As will be illustrated by our experiments, Smart-Inspect can help in the estimation of small-sized defects that cannot be detected by the human eye. Furthermore, our algorithm outperforms state-of-the-arts in differentiating the non-defect regions from the defects. Our contributions are as follows.\\
\begin{itemize}
\item Smart-Inspect can work on raw images (without any enhancement) and full smartphone glass images (without any cropping). This makes it much more powerful and efficient than the current state-of-the-art techniques \cite{li2014defect, liu2011classification}, which can perform only after taking the transparent glass regions by cropping the sides of the glass.
\item Our method outperforms in precise localization and accurate classification of tiny defects (5 microns) that the human eye cannot see.
\item We propose a robust method for the localization and classification of defect and non-defect regions where the system has excellent performance based on a partially labeled small dataset. 
\item Smart-Inspect enables a system for precise labelling of large datasets of smartphone glass with high accuracy. 
\end{itemize}
\section{Formulation and Methodology}
\label{Sec.Formulation and Methodolody}
\subsection{Inspection System Hardware}
\label{sec.inspection}
The experimental setup used in introducing the proposed framework is shown in Fig. \ref{fig1}. A 16K line camera is used to scan an image of the glass. The human eye can detect only a defect over 0.1 millimeters; to decrease this limit, we use a 16K line camera, which captures the defects down to 5 microns. \\
First, the smartphone glass is placed under the camera and a lighting system is mounted at a certain height, as shown in Fig. \ref{fig1}. While capturing the image, the line camera moves from the top to the bottom of the glass to get a full scan image of the glass. The captured image will pop-up on the LCD screen of our experimental system. A core i9 processor is used in the system for handling the 16K resolution image. The size of each image is approximately 400 MB, which is very large compared to simple camera images. To process such a large-sized image, the Smart-Inspect framework is proposed, which will localize the defects precisely, and classify them efficiently.
\subsection{Proposed Approach}
\label{sec.framework}
Smart-Inspect includes four-stage processing. In stage \uppercase\expandafter{\romannumeral1}, all suspicious white continuous regions on each image are cropped along their bounding boxes, which are detected by the contours finding algorithm \ref{algoa}. In stage \uppercase\expandafter{\romannumeral2}, a pre-trained convolutional neural network (CNN) \cite{he2016deep} is used to extract the features of those crops. \\
\begin{figure}
\includegraphics[width=\linewidth]{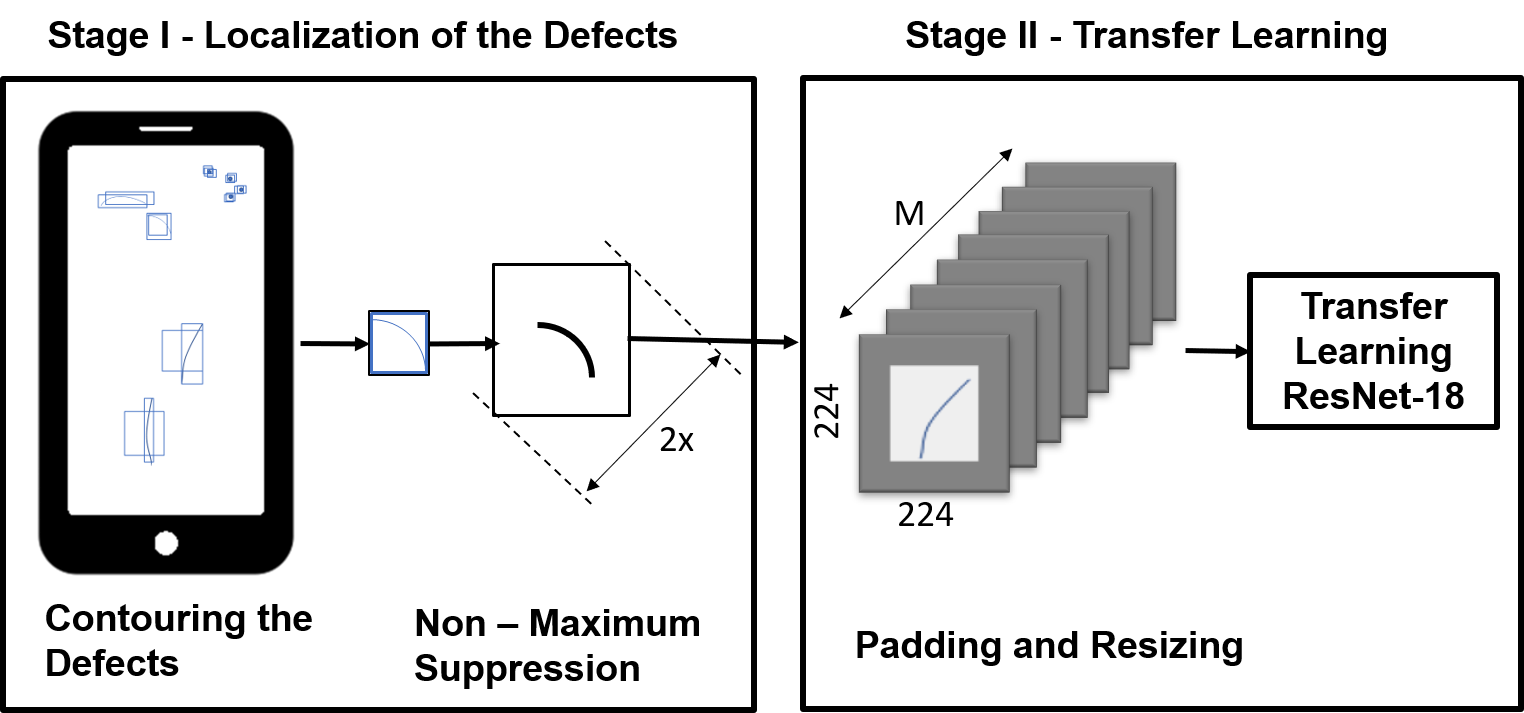}
\centering
\caption{Non-maximum suppression is applied to all the regions. We pad each sample by zeros before the feature extraction for the transfer learning using ResNet-18.}
\label{fig4}
\end{figure}
In stage \uppercase\expandafter{\romannumeral3}, based on the characteristics of our dataset, we have only a few labeled defects and non-defect light reflections due to dust and at the sensors regions. A background/defects (\uppercase\expandafter{BD}) classifier is a binary classifier that divides the top proposals regions into two classes: defects and non-defect areas. It also controls the feedback iterations from unlabeled-proposals to the K-means clusters. We use K-means several times and drop those clusters excluding or containing a relatively low proportion of labeled defects in each loop until the number of dropped crops is less than the preset threshold. This stage cuts the redundant non-defect area crops and greatly reduces the actual number of defects. 

\begin{algorithm}
\KwData{Glass Image $I$ of size $[16384 \times 24576]$ captured from the 16K line camera.\\
    $T_{nms}$ is the non-maximum suppression threshold}
\KwResult{$R = \{r_1,...,r_M\}$ continuous regions boxes of size $[224 \times 224]$ } 
initialization\;
$I_s = {Sobel}_{Operator}(I)$ using kernel $(5,5)$\;
$ I_b = {Threshold}_{binary}(I_s)$ at value of $200$\;
$ I_{dilated} = Dilation(I_b)$ using kernel $(3,3)$\;
Contours $C = \{c_1,...,c_N\}$ of all the white regions proposals (pixels) with detection score $S = \{s_1,...,s_N\}$\;
$R \leftarrow \{\} $\;
$T_{nms} = 0.2$ \;
\While{$C \neq empty $  }{
    $i \leftarrow argmax \: S$ \;
    $O \leftarrow c_i$ \; 
    $R \leftarrow R \bigcup O$ \;
    $C \leftarrow C - O$ \;
    \For{$c_j$ in $C$}{
        \If{iou($O$, $c_j$) $\geq$ $T_{nms}$ }{$ C \leftarrow C-c_j$\;   $S \leftarrow S - s_j$ \; }
    }
}
\While{$R \neq empty $  }{
    $[w,h] = size(R_i)$ \;
    \If{$w < h$ }{
        $R_i \leftarrow padding_{zeros}^{x=h}(R_i) $\;   }
    \ElseIf { $w > h$}{
        $R_i \leftarrow padding_{zeros}^{y=w}(R_i) $  \;  }
    $R_i(224,224) \leftarrow R_i$ \;
}
return $R$
\caption{Continuous Regions Selection Algorithm}
\label{algoa}
\end{algorithm}
In the final stage, stage \uppercase\expandafter{\romannumeral4}, a random forest (RF) is utilized. A six-class defects classifier (\uppercase\expandafter{DC}) based on RF, is trained by the labeled defects. \\

The four-stage processing works in three sections;
\begin{figure*}[ht]
\centering
\includegraphics[width=.90\linewidth]{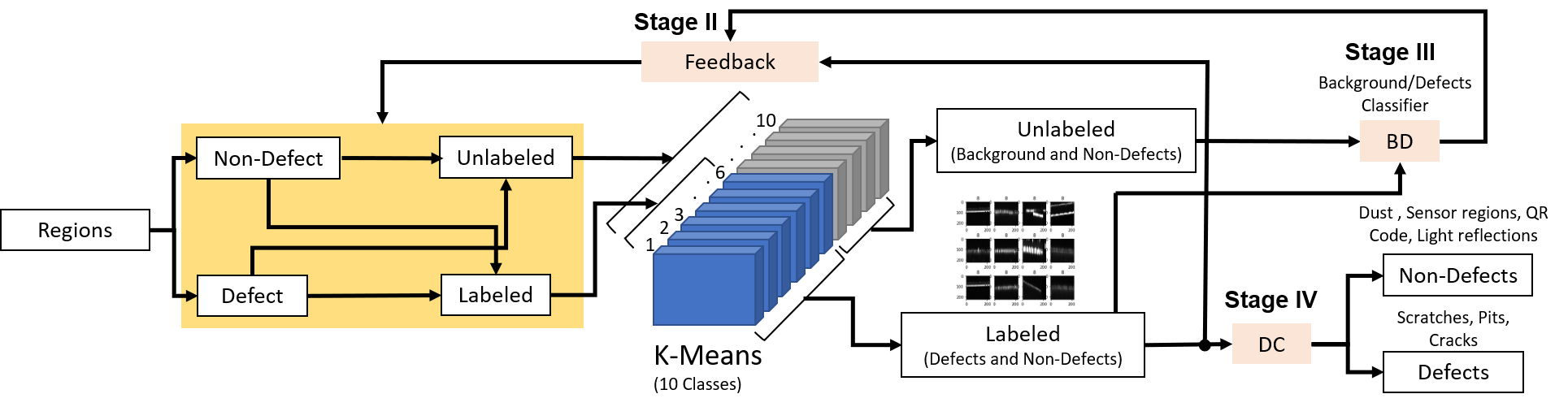}
\caption{Semi-supervised learning for defects and non-defect area classifications. We partially label some regions as the defects type and the non-defect areas.}
\label{fig5}
\end{figure*}
\subsubsection{Dataset}
Our dataset consists of 274 glass images. All of the 16K pixel images have different combinations of defects and non-defect regions.\\
\subsubsection{Localization of the defects}
\label{b2}
Firstly a Sobel operator with a kernel size of 5 is applied to the grey image of the glass. Then, this image is converted into a binary image and dilation is applied on the binary image to enhance the connectivity of the regions. After this, we estimate the bounding box along with its size and original position for each continuous white region by non-maximum suppression \cite{bodla2017soft}. Intersection over Union (IoU) is estimated between each bounding box with the other bounding boxes. Based on the non-maximum suppression threshold ($T_{nms}$), these boxes are merged into one. This scheme is further explained in algorithm \ref{algoa}. $O$ denotes the boxes with maximum confidence. All of the filtered continuous white regions $R$ are padded by zeros to achieve square regions resulting in a single class object. Finally, all the square boxes that correspond to a single class object are resized to $224 \times 224$ pixel batches. For the entire dataset, we obtained $ R = 226,222 $ crops in total, as shown in Fig \ref{fig4}.\\
\subsubsection{Distinguish between the real defects}
All the top proposals include different types of regions such as sensor regions, dust particles, QR code, scratches, dust, pits, cracks, and fingerprints regions. As our approach is semi-supervised, we label 1070 crops manually. Non-defect areas because of light reflections, scratches, pits, cracks, dust and sensor regions including QR codes consist of 30, 270, 210, 280, 150 and 130 regions, respectively. Transfer learning is used to process all the crops for the classification. The pre-trained neural network ResNet-18 \cite{he2016deep} is used as a feature extractor. Each crop is converted to a 512-dim feature vector after passing through ResNet-18.\\
\subsubsection{K-means Clustering}
\label{Sec.K-means}
In order to distinguish real defects from non-defect areas, we use an RF to train the BD classifier. Due to the limited number of labeled data, we cannot train the BD classifier directly by using the supervised approach. Therefore, we design a semi-supervised method to train this classifier. We use k-means iteratively for assigning data to non-overlapping subgroups (clusters), a type of unsupervised learning followed by an approach known as expectation-maximization (EM). Afterwards, we filter some clusters based on the refined labels until this algorithm converges to the optimum value. The loss function used for k-means clustering is as follows:
\begin{align*}
J = \sum_{i=1}^{m} \sum_{k=1}^{K}  w_{ik} \norm{x^i - \mu k} ^ 2,
\end{align*}
For each k-means iteration, all data points (the feature vector of each crop) are divided into 10 clusters (K=10). Then, based on the labelled data, we retain the top six clusters containing the highest proportion of labeled data and cast them into the next iteration, specifically, by using the feedback, as shown in Fig. \ref{fig5}. The iteration continues until the number of dropped data is less than the preset threshold. After dropping all non-defect area data, we retain real defects. We train an intra-class classifier DC to classify the real defects into three different defects: scratches, pits, and cracks, along with three non-defect area objects: dust, sensors regions and on-glass light reflections.\\
\section{Experiments and Results}
\label{Sec:Experiments}
We evaluate the robustness of the Smart-Inspect scheme on glass images with different combinations of defects, as well as on some positive marked glass samples.
\begin{table*}[t]

    \caption{Accuracy Evaluation of the Proposed Method}
    \centering
    
    \begin{tabular}{@{}lcccccccccccc@{}}
    \toprule
    Evaluation Glasses & Defected Regions & Type & TP & FN &    TN &    FP &    Sensitivity &   Specificity & Precision    & Overall Accuracy    \\
    \midrule
    Clean Sample            & 133        & D, LL, SR                     & 128                                     & 3            & 1&0 &0.9770 & \textbf{1}& \textbf{1}&\textbf{ 0.9777 }       \\
    Dust Sample               &     235 &  D, LL, SR                            & 177                                      &     0            & 0&58 &1 &0 &0.7531 &   0.7531         \\
    Scratch Sample                &         191   &  S, D, LL, SR                & 181                                     &    0            & 9 & 1  & 1& 0.9 &\textbf{0.9945} &     \textbf{0.9947}  \\
    Pit+Crack Sample                   &       134      & P, D, LL, SR                   & 126                                      &     0            & 1 & 7 & 1&0.125 &\textbf{ 0.9473}& \textbf{ 0.9477  }      
     \\ \bottomrule
    \end{tabular} %
    \label{tab:2}
    \end{table*}

\subsection{Quantitative Evaluation}
\label{sec:quanti-eva}
Table \ref{tab:2} shows the quantitative evaluation of our semi-supervised method for the glass inspection scheme. Where TP, FN, TN, and FP corresponds to true positive, false negative, true negative, and false positive, respectively. All the test images have combinations of defects on them, such as scratches (S-type), pits (P-type), and light leakage (LL-type) due to non-defect regions of dust (D-type) and sensors regions (SR-type). Our system outperformed on all samples having scratches, pits and cracks. We observed that if the system is not placed in a dust-free environment, then the dust light leakage is a component on all the samples. For the samples with dust, the overall accuracy significantly decreased to 75\% due to the wrong prediction of dust regions as pits, scratches and cracks.
\subsection{Qualitative Evaluation}
\label{sec:qualit-eva}
\begin{figure}
\includegraphics[width=6cm]{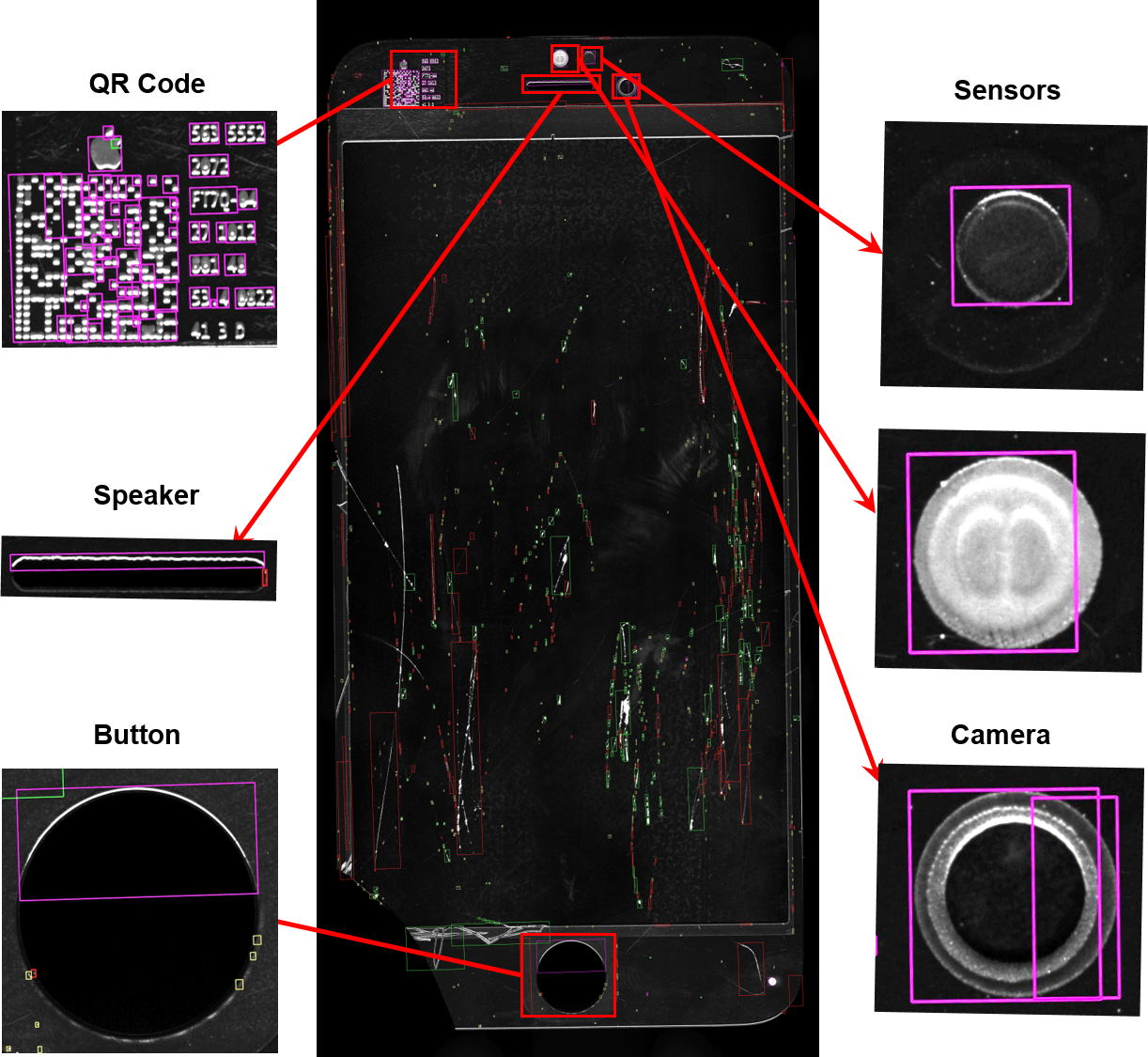}
\centering
\caption{Smart-Inspect test performance on key-challenging test images with hundreds of defects. Sensors regions such as the QR code, camera, and speaker button are successfully distinguished from the defects.}
\label{fig6}
\end{figure}

\begin{figure}
\centering
\begin{subfigure}[t]{0.48\textwidth}
\centering
\includegraphics[width=0.8\textwidth]{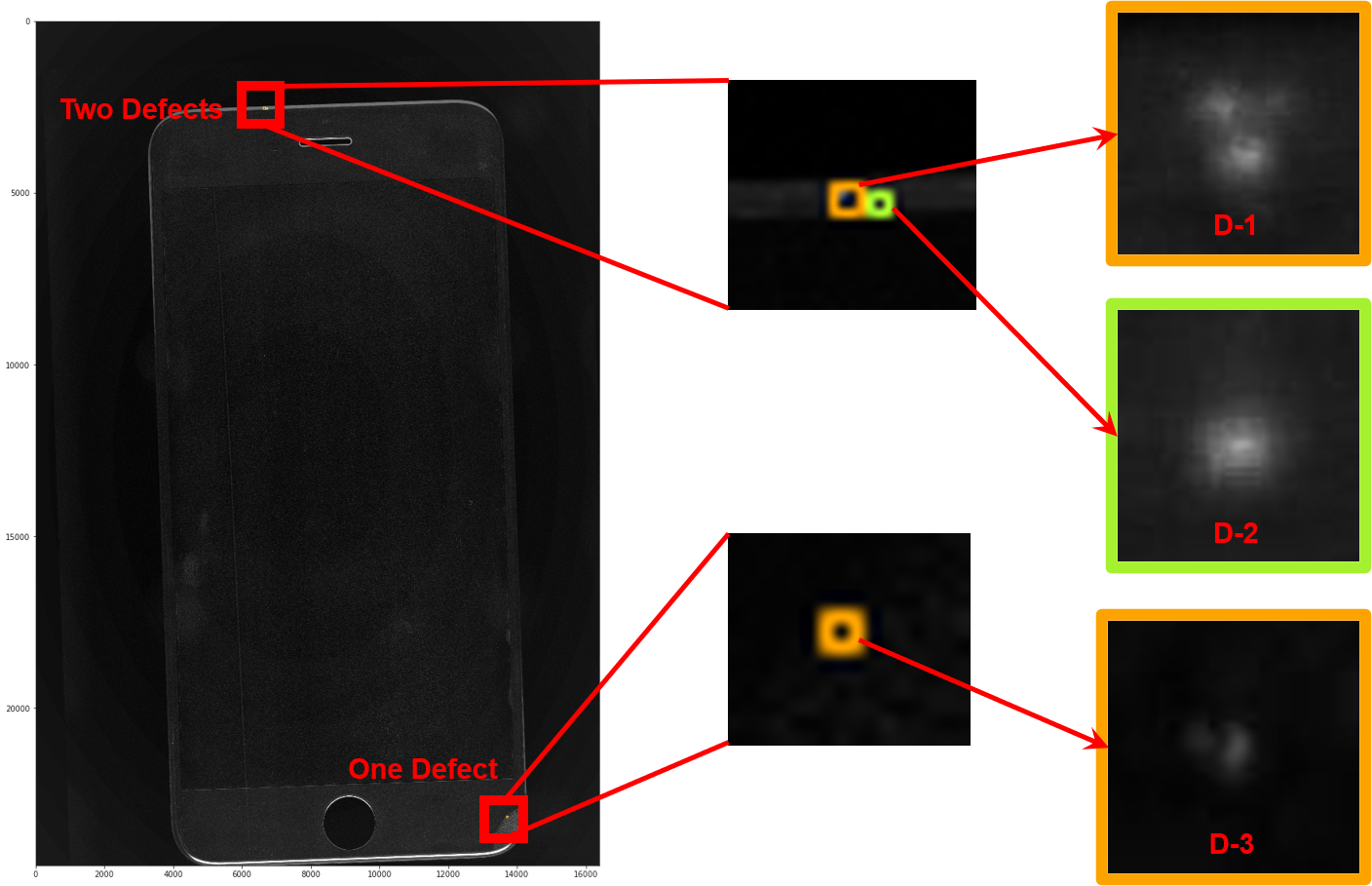}
\caption{Orange boxes (D-1) and (D-3) shows the detected dust on the glass screen. Light leakages due to pits are shown in the green box (D-2).}
\label{fig7a}
\end{subfigure}
\begin{subfigure}[t]{0.45\textwidth}
\centering
\includegraphics[width=0.8\textwidth]{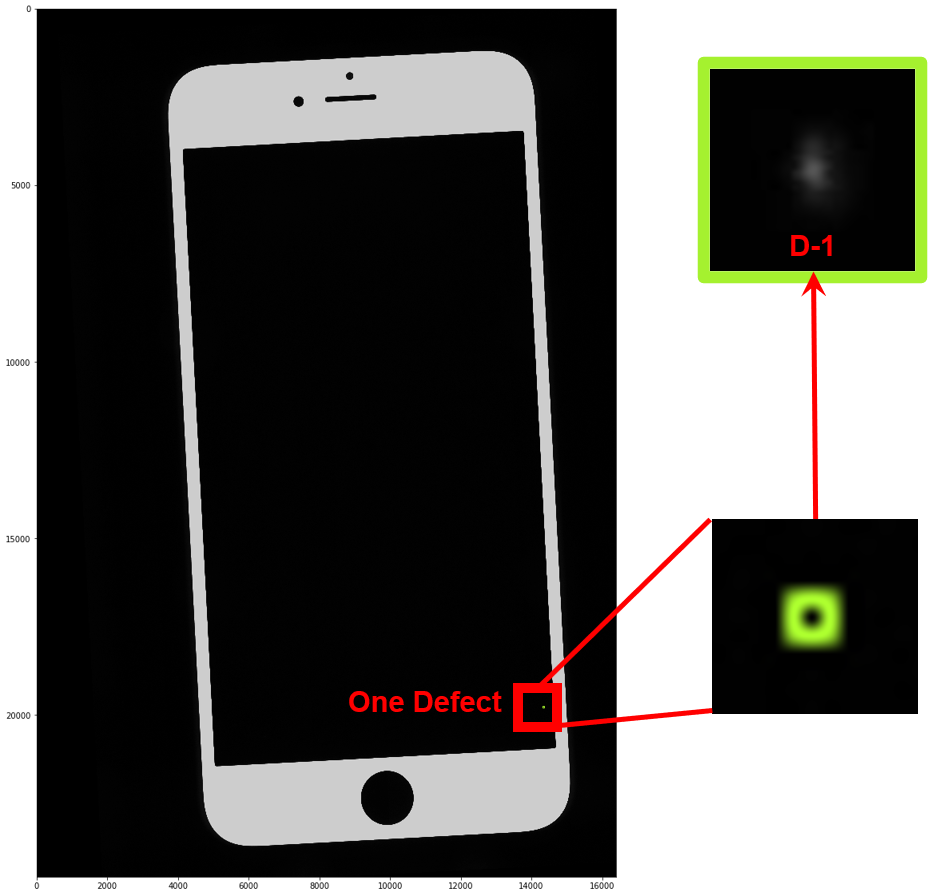}
\caption{One pit (D-1) is observed at the bottom side of the glass.}
\label{fig7b}
\end{subfigure}
\caption{Smart-Inspect performance on the human-marked positive samples. The green box shows light leakage due to a pit, while the orange box express dust region.}
\label{fig7}
\end{figure}

Fig. \ref{fig6} shows the performance of our Smart-Inspect algorithm by inspecting an image of glass with an immense number of defects. We separated the defects using bounding boxes. Regions marked in red are scratches, green represents pits and cracks, yellow shows light leakages due to dust particles. Light leakages from the sensor regions are marked in purple. For our dataset, we chose the RF method, which performed better than support vector machine (SVM) in generating the six classes classifier by training 1072 labeled pieces of data.\\
We observed an interesting fact that Smart-Inspect scanned the entire image and intelligently identified the non-defect regions. Frameworks \cite{li2014defect} and \cite{liu2011classification} treated all the white regions as defects. However, Smart-Inspect did not mark the non-defect regions as defects. Fig. \ref{fig6} clearly shows the non-defect regions such as the QR code, speaker, button, sensors, and camera, which are displayed in purple bounding boxes. Light reflections due to dust particles are shown in yellow boxes in Fig. \ref{fig6}.\\
Furthermore, we tested our algorithm on glass samples that have been previously marked as positive by human inspection. Fig. \ref{fig7} shows two different glass inspection results of positive samples. We notice that Smart-Inspect outperformed human inspection in evaluating the quality of the glass samples. Three regions were localized and classified: two light leakages due to dust particles (D-1 and D-3) and one pit (D-2), which are shown in orange and green boxes, respectively in Fig \ref{fig7a}.\\
In Fig \ref{fig7b}, light leakage due to a pit is detected on the screen area of the smartphone glass. It is successfully marked as a defect region, yet would have been treated as non-defect glass by human inspection. The dot is visible due to the marked green box, but is otherwise difficult to detect by the naked eye.\\
\section{Discussion}
Table \ref{tab:1} shows the comparison of the performance of Smart-Inspect with that of the current advanced methods. It shows that our system is quite intelligent in recognizing non-defect areas. It is also robust to achieve a high level of accuracy of down to 5 micron defects, which is a very important factor of smart device glass quality. \\
Using Smart-Inspect, we can label many 16K smartphone glass images, which are considered challenging to correctly label on a micro scale. Smart-Inspect can assist in the generation of ground truth images much faster than doing so manually, as it helps in the precise cropping and classification. Once all the classes are refined, it is easy to generate thousands of images with labeled defects. Another method is putting defect patches on the smartphone glass and registering the labels for the ground truth.
\subsection{Ablation Study}
Entire image of the glass can be taken as input without enhancing the image defects by applying filtration tools and increasing the sharpness of the captured images, our proposed technique outperformed the state-of-the-arts on all the samples. Note that the images of the glass shown in Fig. \ref{fig6}-\ref{fig7} are enhanced for better visualization in this paper. 
By using a 16K camera, we are able to detect defects down to 5 microns. Although it also detects dust and shows good results, the capturing system requires further improvements, such as a dust-free, controlled working environment. Because dust can significantly affect the results, as explained in Table \ref{tab:2}. Sometimes specks of dust are predicted as a scratch or pit by the system. Therefore, the sample must be put in an experimental setup using gloves to avoid introducing contamination, such as fingerprints, on the screen.
\subsection{Minimization of Inspection Time}
Reducing the inspection time is another key-challenging problem in defect detection. Several methods exist to localize objects in real-time. Faster R-CNN \cite{ren2015faster} is one of the best frameworks; However, it requires a vast labeled dataset. Smart-Inspect now makes it possible to label objects over smartphone glasses to be used with Faster R-CNN for real time defect detection. Processing each sample currently takes about 10-20 sec, which is quite long for industrial applications. Online network systems, where many computers are connected with the server \cite{liu2011classification} may help in enhancing the throughput of the overall system and this will be further enhanced using GPU integration.
\begin{table}[]
\renewcommand{\arraystretch}{1.5}
\caption{Robustness Comparison of the Proposed Method}
\centering
\resizebox{0.5\textwidth}{!}{%
\begin{tabular}{@{}lccc@{}}
\toprule
Inspection Feature              & 
\begin{tabular}[c]{@{}c@{}} PCA based method  \\ \cite{li2014defect} \end{tabular} &  \begin{tabular}[c]{@{}c@{}} MIF Based method  \\ \cite{liu2011classification} \end{tabular} & \begin{tabular}[c]{@{}c@{}} Proposed Method  \\ (Smart-Inspect) \end{tabular} \\   \midrule
Detection of defects                                                          & YES    & YES             & YES    \\
Classification of defects                                                      & NO    & YES             & YES    \\
\begin{tabular}[c]{@{}l@{}}Distinguishing defects \\ and non-defect area\end{tabular} & NO    & NO             & YES    \\ \bottomrule
\end{tabular}%
}
\label{tab:1}
\end{table}
\section{Conclusion} 
\label{sec:conclusion}
According to the experimental results, our semi-supervised method shows excellent performance with high accuracy on a micro scale. It is capable of processing the smartphone glass image as a whole without cropping the transparent region from it. Our approach has the ability to meet the high demand of quality-inspection in production lines of several smart devices in order to compete in the market. Furthermore, the current localization time of the defects can be reduced by labeling vast smartphone glass images using Smart-Inspect.   \\
\bibliographystyle{Ieeetran}
\bibliography{root}

\end{document}